\documentclass[runningheads]{llncs}

\usepackage{graphicx}
\usepackage[bookmarksopen=true,bookmarks=true]{hyperref}
\usepackage{bookmark}

\usepackage[misc]{ifsym}
\usepackage[disable]{todonotes}
\usepackage{lipsum}

\usepackage{amsmath}
\usepackage{amssymb}
\usepackage{mathtools}
\usepackage{bm}
\usepackage{bbm}
\usepackage{mathtools}
\usepackage{cuted}

\usepackage{apptools}
\usepackage{ntheorem}

\newcommand{\Q}{\mathbb{Q}}

\newcommand{\deltamin}{\Delta_{\text{min}}}
\newcommand{\lambdamin}{\lambda_{\text{min}}}
\renewcommand{\P}{\mathbb{P}}

\newcommand{\eps}{\varepsilon}
\newcommand{\cC}{\mathcal{C}}
\newcommand{\cH}{\mathcal{H}}

\newcommand{\egram}{\hat{{\bm G}}}
\newcommand{\cegram}{\hat{{\bm M}}}
\newcommand{\bgram}{\bar{\bm G}}
\newcommand{\emoy}{\overline{\Phi}}
\newcommand{\alphasoft}{\hat{\alpha}_{\mathrm{soft}}}

\newcommand{\minortho}{\underset{\begin{subarray}{l}\|u\|=1\\\bm{1}^Tu=0\end{subarray}}{\text{min}}}

\newcommand{\verticalspace}{\vphantom{\hat{\mathbb{Q}})}}

\AtAppendix{\counterwithin{lemma}{section}}
\AtAppendix{\counterwithin{corollary}{section}}
\AtAppendix{\counterwithin{theorem}{section}}
\AtAppendix{\counterwithin{definition}{section}}
\AtAppendix{\counterwithin{remark}{section}}

\toctitle{Label Shift Quantification with Robustness Guarantees via Distribution Feature Matching}
\tocauthor{Bastien~Dussap}

\begin{document}
\title{Label Shift Quantification with Robustness Guarantees via Distribution Feature Matching}
\titlerunning{Label Shift Quantification with Robust Guarantees}

\author{Bastien Dussap\inst{1}\Letter \and
Gilles Blanchard\inst{1} \and
Badr-Eddine Chérief-Abdellatif\inst{2}
\email{bastien.dussap@univerite-paris-saclay.fr \and
gilles.blanchard@univerite-paris-saclay.fr \and
badr-eddine.cherief-abdellatif@cnrs.fr}}

\authorrunning{B. Dussap et al.}

\institute{Université Paris-Saclay, CNRS, Inria, Laboratoire de mathématiques d’Orsay \and
CNRS, LPSM, Sorbonne Université, Université Paris Cité, France}

\maketitle              

\begin{abstract}

Quantification learning deals with the task of estimating the target label distribution under label shift. In this paper, we first present a unifying framework, distribution feature matching (DFM), that recovers as particular instances various estimators introduced in previous literature. We derive a general performance bound for DFM procedures, improving in several key aspects upon previous bounds derived in particular cases. We then extend this analysis to study robustness of DFM procedures in the misspecified setting under departure from the exact label shift hypothesis, in particular in the case of contamination of the target by an unknown distribution. These theoretical findings are confirmed by a detailed numerical study on simulated and real-world datasets. We also introduce an efficient, scalable and robust version of kernel-based DFM using Random Fourier Features.

\end{abstract}

\section{Introduction}\label{sec:intro}

The success of supervised learning over the last decades is mainly based on the belief that the training and test samples follow the same data generation process. However, in real-world applications, this assumption is often violated and classical learning methods are challenged. Unsupervised domain adaptation, a field of transfer learning, specifically addresses this problem by transferring knowledge from the different but related \textit{training} or \textit{source} domain to the \textit{test} or \textit{target} domain of interest \cite{quinonero2008dataset,patel2015visual}.

From a formal point of view, consider a covariate space $\mathcal{X}$, typically a subset of $\mathbb{R}^d$, and a label space $\mathcal{Y}=[c] := \{1,\ldots,c\}$. We define the two \textit{source} and \textit{target} domains as different probability distributions over the covariate-label space pair $\mathcal{X}\times\mathcal{Y}$. The target label distribution is denoted $\alpha^*=(\alpha_i^*)_{i=1}^c$ while each class-$i$ conditional target distribution is denoted $\Q_i$. Similarly, the source label distribution is denoted $\beta^*=(\beta_i^*)_{i=1}^c$ while each class-$i$ conditional source distribution is denoted $\P_i$. 
We will consider the classical label shift hypothesis:
\begin{equation}
\label{cond:LS}
\tag{$\mathcal{LS}$}
\forall i=1,\ldots,c \hspace{0.1cm} , \quad \P_i=\Q_i.
\end{equation}

Another setting we will consider involves contamination of the target by a new class. In this \textit{contaminated label shift} setting, we assume that the target domain is $\mathcal{X}\times\tilde{\mathcal{Y}}$, with $\tilde{\mathcal{Y}}= \{0,\ldots,c\}$ and that the label shift hypothesis is still verified for the class $\{1,\ldots,c\}$: 
\begin{align*}
\label{cond:CLS}
\tag{$\mathcal{CLS}$}
&\Q = \sum_{i=1}^c \alpha_i^* \P_i + \alpha_0^*\Q_0 \\
&\forall i=1,\ldots,c \hspace{0.1cm} , \quad \P_i=\Q_i.
\end{align*}
The distribution $\Q_0$ is seen as a noise or a contamination, for which we have no prior knowledge nor sample. Therefore, our objective in this contaminated scenario is to be robust to a large class of noise distributions. In Section \ref{sub:mis}, we will give insight on the kind of contamination we can be robust to.

In both settings, we suppose a source dataset $\{(x_j,y_j)\}_{j\in[n]}\in\left(\mathcal{X}\times\mathcal{Y}\right)^n$ and a target dataset $\{x_{n+j}\}_{j \in [m]} \in\mathcal{X}^m$ are given. All data points from the source (respectively the target) dataset are independently sampled from the source (resp. the target) domain. We have access to the source labels $y_j$ but not to the target labels which are not observed.
We denote by $\hat{\P}_i:=\sum_{j \in [n]:y_j=i} \delta_{x_j}(\cdot)/n_i$ the empirical source class-$i$ conditional distribution, where $\delta_{x_j}$ denotes the Dirac measure at point $x_j$ and $n_i$ the number of instances labeled $i$ in the source dataset. Note that $n_1+\ldots+n_c=n$. We finally denote by $\tilde{\beta}$ the empirical proportions of each class in the source dataset, i.e. $\tilde{\beta_i} := n_i/n$.

Several different objectives have been addressed under the label shift assumption in the literature, and can be summarised in three points: (i) \textit{detection}, i.e. determining whether distribution shift has occurred; (ii) \textit{correction}, i.e. fitting a classifier with high accuracy on the target distribution; and (iii) \textit{quantification}, i.e. estimating the target label distribution \cite{saerens2002adjusting,forman2005counting,gonzalez2017review,lipton2018detecting,azizzadenesheli2019regularized,garg2020unified}. We focus here on the last challenge, and develop a general analysis unifying several existing techniques with estimation guarantees for the target proportions $\alpha^*$, as well as dealing with the contaminated label shift setting.

\subsection{Related literature} \label{sub:related}
The research area of quantification learning has a somewhat fragmented nature.
Quantifying the target label distribution and learning a classifier are
actually very closely related objectives. The most classical approach for the construction of efficient classifiers on the target domain is based on weighted empirical risk minimisation, which itself requires the estimation of the shift between the source and target distributions. Thus, while we are here interested in estimating the target proportions $\alpha_i^*$, many related works are interested in estimating the weights $w_i = \alpha_i^*/\beta_i^*$. Obtaining an estimator of those weights from an estimator of the target proportions $\alpha_i^*$ is straightforward: simply use the labels in the source data to form a direct estimate of the source proportions and then consider the ratio. Conversely, starting from the weights estimator, it is possible to obtain an estimate of the weights $\alpha_i^*/\beta_i^*$ by multiplying it by an estimate of the source proportions. For this reason, there have been two different literature threads that address closely related problems but apparently grew independently: the Quantification Learning literature \cite{gonzalez2017review,esuli2023learning} and the Label Shift literature \cite{garg2020unified}.

Most methods dealing with label shift are expressed as variants of the so-called \textit{Classify \& Count (CC)} technique proposed in the seminal works of Forman \cite{forman2005counting,forman2006quantifying,forman2008quantifying}. The idea is to fit a classifier on the source dataset, e.g. a random forest \cite{milli2013quantification} , an SVM \cite{barranquero2015quantification}, or a Nearest-Neighbour \cite{barranquero2013study}, and to estimate the target class distribution using the distribution of predictions. To account for the misclassification that the underlying classifier suffers on the target set due to the label shift, Forman \cite{forman2008quantifying} proposed the \textit{Adjusted Classify \& Count (ACC)} method, a modification to the standard Classify \& Count which simply adjusts the estimate after computing the quantifier. This approach is also popular in the label shift correction literature. Based on the same principle, more
recently the \textit{Black-Box Shift Estimation (BBSE)} algorithm introduced by \cite{lipton2018detecting} used the confusion matrix of the classifier to adjust the predicted label distribution, while Azizzadenesheli et al. \cite{azizzadenesheli2019regularized} proposed to regularise the re-weighting vector in BBSE in order to make the final estimated target distribution less sensitive to the quality of the estimated confusion matrix. Another technique, using an off-the-shelf classifier, is based on the maximum likelihood principle. The most popular version of this approach is probably the \textit{Maximum Likelihood Label Shift (MLLS)} \cite{alexandari2020maximum} strategy which is widely used in the label shift correction community. This technique is actually a variation of the original work of Saerens et al. \cite{saerens2002adjusting} that uses an Expectation-Maximisation (EM) algorithm to efficiently correct for the shift in class proportions between source and target distributions given estimates from a predictive model, alternately updating the target class-prior and class-posterior probabilities from some initial estimates until convergence. In fact, \cite{garg2020unified} argues that only the choice of the calibration method differs between MLLS and BBSE and that both procedures actually solve the same optimisation objective, thus explaining that the empirical advantage of MLLS over BBSE is only due to coarse calibration of the latter. 

Another completely different approach consists in viewing quantification as a statistical mixture problem. Since both source and target covariates marginal distributions  can be written as mixtures $\P = \sum_{i=1}^c \beta_i^* \P_i$ and $\Q = \sum_{i=1}^c \alpha_i^* \Q_i$ respectively. Under the label shift assumption, the conditional distribution of the covariates given the label is the same for both source and target data ($\P_i=\Q_i$), and can be estimated using the empirical conditional distribution $\hat{\P}_i:=\sum_{j/y_j=i} \delta_{x_j}(\cdot)/n_i$ based on the labeled source sample. Thus, the marginal covariates distribution $\Q$ can be approximated by $\sum_{i=1}^c \alpha_i^* \hat\P_i$, and the goal of quantification can be seen as finding the mixture weights $(\alpha_i)_{i=1}^c$ such that the mixture $\sum_{i=1}^c \alpha_i \hat\P_i$ resembles the most the empirical target covariates marginal distribution $\hat{\Q}:=\sum_{j=1}^m \delta_{x_{j+n}}(\cdot)/m$ based on the target dataset, with respect to some metric. Many different statistical divergences have been considered in the literature, such as the Hellinger distance \cite{gonzalez2013class}, the Wasserstein distance \cite{bigot2022potential}, the Pearson divergence \cite{du2014semi},
the Energy distance \cite{kawakubo2016computationally} or the Maximum Mean Discrepancy (MMD) \cite{iyer2014maximum,zhang2013domain}. The last two distances operate in a Reproducing Kernel Hilbert Space (RKHS) and adapt to the label shift problem the Kernel Mean Matching (KMM) approach \cite{gretton2009covariate}, that minimises the RKHS distance between the kernel mean embeddings of the distributions. The use of kernel methods enables here to operate in an implicit high-dimensional space without requiring computing the coordinates of the data in that space, but rather by simply computing the inner products between the images of all pairs of data in the space. However, the applicability of these methods on a larger scale still remains an important limitation due to the computation of the kernel matrix. Consequently, distribution matching has sometimes been performed on a low-dimensional function of the covariates instead of a high-dimensional kernel embedding. For instance, the Distribution y-Similarity framework (DyS) \cite{maletzke2019dys} exploits the histogram of a decision function obtained by a trained binary classifier and then minimises the Tops{\o}e distance between the two histograms, while the HDx algorithm \cite{gonzalez2013class} uses the Hellinger distance between histograms of the Source and Target directly.  \textit{The general formulation we will adopt in this paper is in line with this approach of minimising some distance in a feature space between feature mappings of the distributions.} 

\subsection{Contributions of the paper}\label{sub:orga}

We introduce a general framework for label shift quantification, \textit{Distribution Feature Matching} (DFM),  that generalises existing methods such as Black-Box Shift Estimation (BBSE) \cite{lipton2018detecting}, Kernel Mean Matching (KMM) \cite{iyer2014maximum,zhang2013domain} and its variant Energy Distance Matching \cite{kawakubo2016computationally}.

The contributions of the paper are the following:
\begin{enumerate}
    \item In Section \ref{sec:dfm}, we propose a general framework of Label-shift quantification based on the minimisation of the distance between representations of distributions (in a Euclidean or Hilbert space) obtained by taking expectations of a feature mapping. We show that existing methods in the literature such as KMM or BBSE are particular instances of this general framework.
    
    \item In Section \ref{sec:theory}, we provide a general statistical analysis of DFM under label shift. In particular, we show that our bound on the estimation error significantly improves the previous ones derived in the literature for both KMM and BBSE. 

    \item We also derive a novel analysis of DFM methods under departures of the label shift hypothesis, using a geometric decomposition of the problem and we show the implication of this analysis when we are in the contamination setting presented earlier. We thus show that certain DFM methods can exhibit robustness against particular types of perturbations or contaminations.
   
   \item In Section \ref{sec:appli}, we support our theoretical results regarding the robustness of the different methods, with experiments on synthetic Gaussian mixtures and real cytometric datasets.

  \item Finally, we implement the KMM procedure in Python, using fast GPU-compatible code, using Random Fourier Features (RFF) to significantly reduce the computation burden while keeping the same type of theoretical guarantees. This implementation can still be used on GPU with limited memory. 

\end{enumerate}

\section{Distribution Feature Matching} \label{sec:dfm}

Let $\Phi:\mathcal{X} \rightarrow \mathcal{F}$
be a fixed feature mapping from $\mathcal{X}$ into
a Hilbert space $\mathcal{F}$
(possibly $\mathcal{F}=\mathbb{R}^D$).
We extend the mapping $\Phi$
to probability distributions on $\mathcal{X}$ via taking expectation, i.e. 
$\Phi \colon \P \mapsto \Phi(\P) := \mathbb{E}_{X\sim\mathbb{P}}[\Phi(X)] \in \mathcal{F}$. Thus, it holds $\Phi(\hat{\mathbb{P}}_i) = n_i^{-1} \sum_{j \in [n]: y_j=i} \Phi(x_j)$, and similarly $\Phi(\hat{\mathbb{Q}}) = m^{-1} \sum_{j=n+1}^{n+m} \Phi(x_j)$.
 
We call \textit{Distribution Feature Matching} (DFM) any estimation procedure that can be formulated as the minimiser of the following problem:
\begin{equation}
\label{P:DFM}
\tag{$\mathcal{P}$}
\hat{\alpha} = \underset{\alpha \in \Delta^c}{\arg \min}\  \left\| \sum_{i=1}^c \alpha_i \Phi(\hat{\mathbb{P}}_i)-\Phi(\hat{\mathbb{Q}}) \right\|_{\mathcal{F}}^2
\end{equation}
where $\Delta^c$ is the $(c-1)$-dimensional simplex.\\
In the contamination setting, we aim at finding the proportions of the non-noise classes of the target. As these proportions don't sum to one, the "hard" condition $\sum_{i}\alpha_i = 1$ is no longer needed. One way to overcome this is to introduce a fictitious "dummy" class in the source that formally has a vectorisation equal to $0$ (note that adding a dummy class is a computational and theoretical convenience; we don't require to have a real distribution $\P_0$ that maps to $0$ in the feature space for the results to hold). If we write $\Phi(\hat{\mathbb{P}}_0) := 0$ one can see that:
\begin{equation}
\label{P:DFMsoft}
\tag{$\mathcal{P}_2$}
\underset{\alpha \in \Delta^{c+1}}{\arg \min}\  \left\| \sum_{i=0}^c \alpha_i \Phi(\hat{\mathbb{P}}_i)-\Phi(\hat{\mathbb{Q}}) \right\|_{\mathcal{F}}^2 =
\underset{\alpha \in \text{int}(\Delta^c)}{\arg \min}\  \left\| \sum_{i=1}^c \alpha_i \Phi(\hat{\mathbb{P}}_i)-\Phi(\hat{\mathbb{Q}}) \right\|_{\mathcal{F}}^2, 
\end{equation}
where $\mathrm{int}(\Delta^c) := \{x\in\mathbb{R}^c \colon x\geq 0, \ \sum x_i \leq 1\} $.
A procedure that solves \ref{P:DFMsoft} will be called \textit{soft}-DFM.
In Section \ref{sec:theory}, we will present theoretical results, in the classical Label Shift hypothesis (\ref{cond:LS}), for DFM methods of
the form \eqref{P:DFM}
under an identifiability and boundedness assumption.\\
In Section \ref{sub:mis}, we will show a general result for ("hard") DFM methods when the label shift hypothesis is not verified. 
As a corollary of these bounds we will 
directly obtain corresponding guarantees for the {\em soft}-DFM methods as well through the representation \eqref{P:DFMsoft} and formal inclusion of
the dummy class.

In the remainder of this section, we will show the link between DFM and other classical Label Shift Quantification algorithms. However, any black-box feature mapping will be suitable for the results of Section \ref{sec:theory}. 

\subsection{Kernel Mean Matching}
Iyer et al. \cite{iyer2014maximum} used Kernel Mean embedding 
(KME) as their mapping.
We refer the reader to \cite{muandet2017kernel} for a survey on KME.
We briefly recall that for any symmetric and semi-definite positive kernel $k$ 
defined on $\mathcal{X}$, one can
associate a Hilbert space 
denoted $\mathcal{H}_k$,
or simply $\mathcal{H}$ when there is no ambiguity, and a "feature" mapping $\Phi: \mathcal{X} \rightarrow \mathcal{H}$
such that $\langle \Phi(x),\Phi(y) \rangle = k(x,y)$.

This mapping can be extended to the space of distributions by taking expectations
as described above, which constitutes the principle of KME. 
We can compute scalar products between mappings using the formula
$\langle \Phi(\mathbb{P}), \Phi(\mathbb{Q}) \rangle_{\mathcal{H}} = \mathbb{E}_{(X,Y)  \sim\mathbb{P} \otimes \mathbb{Q}}[k(X,Y)],$
which provides a way to find an explict solution
of~\eqref{P:DFM} in practice.
Then $D_{\Phi}(\mathbb{P}, \mathbb{Q}) = \| \Phi(\mathbb{P}) - \Phi(\mathbb{Q}) \|_{\mathcal{H}}$ is a pseudo-distance on the space of measures on $\mathcal{X}$, called \textit{Maximum Mean Discrepancy} (MMD) \cite{gretton2012kernel}. 
The specific use of a kernel feature mapping for
class proportion estimation via
\eqref{P:DFM} has been called Kernel
Mean Matching (KMM) by \cite{iyer2014maximum}.

If $\mathcal{X} = \mathbb{R}^d$ with the usual
Euclidean norm,
Kawakubo et al. \cite{kawakubo2016computationally} proposed a particular case of KMM using the Energy kernel: $k(x,y) = \|x\| + \|y\| - \|x-y\|$,
which is indeed a reproducing kernel \cite{sejdinovic2013equivalence}.

\subsection{BBSE as Distribution feature matching}

Black-Box Shift Estimation is a method using the output of a black-box classifier to estimate the proportions in the target. To take into account the bias of the training data (i.e. the source) \cite{lipton2018detecting} used the confusion matrix.\\
To understand how Black-Box Shift Estimation can be cast as a Distribution Feature Matching procedure, we start from its original formulation as seeking the vector of proportions $\alpha$ that satisfies $Y={M}\alpha$, where ${M}$ is the estimated conditional confusion matrix defined as $M_{ij} = \frac{1}{n_i} \sum \bm{1}\{f(x_l) = i \ \text{and} \ y_l = j \}$ and $Y$ is the empirical mean of the observed outputs of the black-box classifier $f$ on the target data, $Y_i = \frac{1}{n} \sum_l \bm{1}\{ f(x_l)=i\}$. The BBSE estimate is then $\hat\alpha={M}^{-1}Y$ 
($M$ is explicitly assumed invertible
by \cite{lipton2018detecting}).

\begin{proposition}\label{prop:bbse}
The BBSE estimator based on the black-box classifier $f$ is the same as the solution of the DFM problem \eqref{P:DFM} using the feature mapping 
$\Phi(x) = (\bm{1}\{f(x)=i\})_{i=1,\ldots,c} \in \mathbb{R}^c$, where the positivity constraint on $\alpha$ is dropped.
\end{proposition}
The proof is postponed to appendix \ref{proof:prop:bbse}.

In the experiments to come, we will use BBSE+, our modified version of BBSE including the positivity constraint. The experimental results are slightly better for BBSE+. The reason is that in many cases, due to the presence of small classes in the source and in the target, BBSE returns negative proportions. When it does not output negative values, the two algorithms are the same.\\
This version of BBSE already existed in the literature, it has been used for text content analysis \cite{hopkins2010method} and as a building block for classification, in a domain adaptation setting, more general than Label Shift \cite{tachet2020domain}.

\section{Theoretical guarantees} \label{sec:theory}

We now 
provide statistical guarantees for DFM quantification procedures, 
All proofs can be found in appendix \ref{appendix:prooftheo}. 

We make the following identifiability hypothesis on the mapping $\Phi$:
\begin{equation}
\label{cond:1}
\tag{$\mathcal{A}_1$}
\sum_{i=1}^c \beta_i \Phi(\P_i) = 0 \iff \beta_i = 0 \ \forall i=1,\ldots,c,
\end{equation}
and
\begin{equation}
\label{cond:2}
\tag{$\mathcal{A}_2$}
\exists C>0: \qquad \| \Phi(x) \|_{\mathcal{F}} \leq C \ \text{for all} \ x.
\end{equation}
%
If we use KMM, the boundedness property is satisfied as soon as
the kernel is bounded (e.g. Gaussian kernel, or any continuous kernel on a compact space). For BBSE, the boundedness is verified with $C=1$.

Concerning Condition~\ref{cond:1}, it is satisfied in the KMM case 
as long as the kernel is characteristic (e.g. Gaussian kernel) and the distributions $\P_i$ are linearly independent (which is the minimal assumption for the class proportions to be identifiable). 
These assumption have been previously used by \cite{iyer2014maximum} for KMM. 
Similarly, for BBSE, 
\cite{lipton2018detecting} also assumed identifiability and required that the expected classifier outputs for each class be linearly independent.

We introduce the following notation and state our main theorem:
\begin{definition}
\label{def:gram}
We denote $\egram$ the Gram matrix, resp. $\cegram$ the centered
Gram matrix of $\{ \Phi(\hat{\P}_1), \cdots, \Phi(\hat{\P}_c)\}$.
That is, $\egram_{ij} = \langle \Phi(\hat{\P}_i), \Phi(\hat{\P}_j)\rangle$ and $\cegram_{ij} = \langle \Phi(\hat{\P}_i) - \emoy, \Phi(\hat{\P}_j) - \emoy \rangle$ with $\emoy = c^{-1} \sum_{k=1}^c \Phi(\hat{\P}_k)$. Furthermore, let $\deltamin$ be
the second smallest eivenvalue of $\cegram$ and $\lambdamin$
the smallest eigenvalue of $\egram$. In particular, it holds
$\deltamin \geq \lambdamin$.
\end{definition}

\begin{theorem}
\label{theo:main}
If the Label Shift hypothesis \eqref{cond:LS} holds, and if the mapping $\Phi$ verifies Assumptions \eqref{cond:1} and \eqref{cond:2},
then for any $\delta \in (0,1)$, with probability greater than $1-\delta$, the solution $\hat{\alpha}$ of \eqref{P:DFM} satisfies:
	\begin{align}
	    \| \hat{\alpha} - \alpha^* \|_2 &\leq \frac{2CR_{c/\delta}}{\sqrt{\deltamin}}\left(\frac{\|w\|_2}{\sqrt{n}}  +  \frac{1}{\sqrt{m}}\right) \label{theo:main:eq1} \\
	    &\leq \frac{2CR_{c/\delta}}{\sqrt{\deltamin}}\left(  \frac{1}{\sqrt{\min_i n_i}} +
     \frac{1}{\sqrt{m}}\right), \label{theo:main:eq2}
	\end{align}
where $R_{x} = 2+\sqrt{2\log (2x)}$, $w_i = \frac{\alpha_i^*}{\tilde{\beta_i}}$.

The same result holds when replacing $\alpha^*$
by the (unobserved) vector of empirical proportions
$\tilde{\alpha}$ in the target sample,
both on the left-hand side and in the definition
of $w$.

Under the same assumptions, the solution $\alphasoft$ of \eqref{P:DFMsoft}
satisfies the same bounds with $\deltamin$ replaced by $\lambdamin$.
\end{theorem}


\subsection{Comparison to related literature}

 We compare our result to Theorem 1 of \cite{iyer2014maximum} and Theorem 3 of \cite{lipton2018detecting},
 which as we have mentioned earlier hold under the same
 assumptions as we make here.

Concerning KMM, a comparison between our inequality \eqref{theo:main:eq2} and Theorem~1 in \cite{iyer2014maximum} shows that our bound is tighter than theirs, which is of leading order 
$$
\frac{c}{\sqrt{m}}+\sum_i \frac{1}{\sqrt{n_i}}
\quad 
\text{vs ours in}
\quad
\frac{1}{\sqrt{m}}+\max_i \frac{1}{\sqrt{n_i}} 
$$
up to logarithmic factors. Thus, Theorem \ref{theo:main} improves upon the previous upper bound by a factor of $c$ with respect to the term in $m$, and reduces the sum into a maximum regarding the number of instances per class $n_i$, which may also decrease the order of by factor $c$ when the classes are evenly distributed in the source dataset. Furthermore, Inequality \eqref{theo:main:eq1} even significantly improves over both Inequality \eqref{theo:main:eq2} and Theorem~1 in \cite{iyer2014maximum}. Indeed, in situations where one of the classes $i$ on the source domain is rare, then the rate $\text{max}_i \ n_i^{-1/2}$ in Inequality \eqref{theo:main:eq2} explodes, which is not the case of the rate $ \|w\|_2 /\sqrt{n}$ in Inequality \eqref{theo:main:eq1} when the source and target proportions are similar, as the weight vector $w$ reflects the similarity between the source and target distributions. Note that we use the theoretical proportions $\alpha^*$ for the target in the definition of $w$ as the empirical ones are unknown here. Hence, our theorem significantly improves the existing bound for KMM established by \cite{iyer2014maximum}. Similarly, our bound \eqref{theo:main:eq1} applied to BBSE also improves Theorem 3 in \cite{lipton2018detecting}. In particular, when both inequalities are formulated with the same probability level (e.g. $1-\delta$), our bound for BBSE is tighter by a factor $\sqrt{c}$ w.r.t. the term in $m$ than the guarantee provided by \cite{lipton2018detecting}. Note however that contrary to our result and to Theorem~1 in \cite{iyer2014maximum}, the bound of \cite{lipton2018detecting} does not involve any empirical quantity that can be computed using the source dataset.

Another key component of the bounds is the second smallest eigenvalue $\deltamin$ of the centered Gram matrix, which replaces the minimum singular value of the Gram matrix in the case of KMM (see Theorem~1 in \cite{iyer2014maximum}) and the smallest eigenvalue of the confusion matrix divided by the infinite norm of the source proportions in the case of BBSE (see Inequality (3) of Theorem 3 in \cite{lipton2018detecting}), and improves upon both of them. 

This improvement is particularly important when the two source classes are unbalanced.
For instance, in a two-class setting with a black-box classifier feature map and $\beta^*=(p,1-p)$, the theoretical version of $\deltamin$ is equal to $1$ when the classifier is perfect and replaces the factor that would be $\min\left(\frac{p}{1-p},\frac{1-p}{p}\right)<1$ in the bound of \cite{lipton2018detecting}. When the classifier is not perfect but both classes share the same classification accuracy $a\in(1/2,1)$, then $\deltamin=2a-1$, which strictly improves the factor of \cite{lipton2018detecting} except when both classes are equally balanced, in which case both quantities are equal.

To fully understand the nature of $\deltamin$, 
we can interpret DFM as the projection of the target feature mapping onto the convex hull of the source feature mappings. Our estimation is then simply the barycentric coordinate of the projection, as formalised in \eqref{QP}, and $\deltamin$ is a geometric property of that convex hull. It represents the ease with which our mapping can distinguish between one class and any mixture of the other classes. For instance, for two classes and two embeddings $(\Phi_1,\Phi_2)$, one can show that $\deltamin = \frac{1}{2} \|\Phi_1 - \Phi_2\|_{\mathcal{F}}^2$. From a geometric point of view, it is clear that the larger the convex hull (i.e. the line connecting the two features in situations where there are two classes), the less the barycentric coordinate will be affected by a small perturbation of the weights. From a statistical point of view, if our mixture is composed of two very different distributions, it will be intuitively easier to distinguish them in a new sample. 
The quantity $\deltamin$ (which we recall is empirical) also provides a natural criterion for the choice of the feature map hyperparameter (i.e. choice of the kernel in KMM
), as the dependence in our bound only appears in $\deltamin$ which can then be maximised.

\subsection{Robustness to contamination} \label{sub:mis}

We now introduce a novel theoretical analysis of the robustness of
the method with respect to the contamined label shift
model (assumption \ref{cond:CLS}).
First, let us obtain a general result when Label Shift is not verified.

A naive approach would simply
include the bias term
$\|\Phi(\mathbb{P}_i) - \Phi(\mathbb{Q}_i)\|_{\mathcal{F}}$ in the bound.
We put into light the robustness of the procedure with respect to certain types of perturbation.\\ 
\begin{theorem}
    \label{theo:main_contaminated}
    Denote $V$ the affine span of the vectors $\Phi(\mathbb{P}_i)$ and $\mathcal{C}$ the convex hull of those same vectors. Denote $\Pi_V$ and $\Pi_\mathcal{C}$ the orthogonal resp. convex projection onto $V$ and $\mathcal{C}$.\\
    Suppose the same assumptions as
    in Theorem \ref{theo:main} hold,
    except for the exact label shift assumption
    \ref{cond:LS}.
    Then, with probability greater than $1-\delta$ :
    \begin{equation} \label{eq:robust}
       \| \hat{\alpha} - \alpha^* \|_2 \leq  \frac{1}{\sqrt{\deltamin}}\left(
             3 \epsilon_n + \eps_m 
             + \sqrt{2\epsilon_n} B^\perp + B^\parallel
       \right), 
    \end{equation} with:
 \begin{equation}
     \label{eq:epsilons}
\epsilon_n = C\frac{R_{c/\delta}}{\sqrt{\min_i n_i}}; \qquad
 \eps_m = C\frac{R_{1/\delta}}{\sqrt{m}};
 \end{equation}
 \begin{align*}
    B^\perp = B^\perp(\mathbb{P}, \mathbb{Q}) &= \sqrt{\| \Phi(\mathbb{Q}) - \Pi_{\mathcal{C}}(\Phi(\mathbb{Q})) \|_{\mathcal{F}}};\\
    B^\parallel = B^\parallel(\mathbb{P}, \mathbb{Q}) &= \max_i 
    \| \Phi(\mathbb{P}_i) - \Pi_V( \Phi( \mathbb{Q}_i)) \|_{\mathcal{F}}.
\end{align*}
\end{theorem}
Observe that the bound~\eqref{eq:robust}
shows robustness of a DFM procedure
against perturbations $B^\perp$ that are "orthogonal" to the source space $V$
in feature space.
In particular, {\em consistency} (i.e. convergence of the bound to 0 as the sample sizes grow to infinity) is still granted if $\mathbb{Q}_i \neq \mathbb{P}_i$ but $\Pi_V(\mathbb{Q}_i) = \mathbb{P}_i$.
Which type of perturbation of the distributions
will result in (close to) orthogonal shifts in feature space very much depends on the feature mapping used. For BBSE, the feature space is of the same dimension as the number of sources, thus
under condition \eqref{cond:1}, $V$ will coincide with $E_1$, the affine space of vectors summing to one. Since any distribution will be also mapped to $E_1$,
the orthogonal component will always be 0.
Thus, we expect no particular 
robustness property for
BBSE methods.
For more general feature maps, such as kernel methods or any other vectorisations, this orthogonality property remains to be investigated
in general, but we will exhibit below a 
favorable scenario for KMM in the {\em contaminated label shift} setting \ref{cond:CLS}.

We first state a corollary in the \ref{cond:CLS} scenario. To do so, we recall that, in this case, we use the \textit{soft}-DFM procedure \ref{P:DFMsoft}. 
We are now in a particular case, where the only
difference between source and target
is that the unknown noise class $\Q_0$ is formally replaced by
the dummy class 
having feature map equal to $0$ in the source.
Introduce $\bar{V}:=\mathrm{Span}\{ \Phi(\mathbb{P}_i), i \in [c]\}$ (i.e. vector
span rather than affine span for $V$ previously) and let $\Pi_{\bar{V}}$ be the orthogonal projection on $\bar{V}$.

\begin{corollary}
    \label{cor:main_contaminated}
    Denote by $\alphasoft$ the minimiser of the \textit{soft}-DFM problem \ref{P:DFMsoft}.
    Assume the contaminated Label Shift hypothesis \eqref{cond:CLS} holds, and if the mapping $\Phi$ verifies Assumptions \eqref{cond:1} and \eqref{cond:2}.
    Then, with probability greater than $1 - \delta$:
    \begin{equation*}
       \| \alphasoft - \alpha^* \|_2 \leq  \frac{1}{\sqrt{\lambdamin}}
       \Bigl(
             3 \epsilon_n + \eps_m 
             + \sqrt{2\alpha_0 \ \epsilon_n \ \|\Phi( \mathbb{Q}_0 ) \|_{\mathcal{F}}} + \| \Pi_{\bar{V}}( \Phi( \mathbb{Q}_0)) \|_{\mathcal{F}}
       \Bigl), 
    \end{equation*} with $\epsilon_n,\varepsilon_m$ defined as in~\eqref{eq:epsilons}.
%
\end{corollary}

In the particular case of KMM with a translation-invariant
kernel $k(x,y)=\varphi(x-y)$, 
for any distributions $\mathbb{P},\mathbb{P}'$ it holds
$\langle \Phi(\mathbb{P}),\Phi({\mathbb{P}'}) \rangle = 
\mathbb{E}_{(X,Y) \sim \mathbb{P} \otimes \mathbb{P}' }[\varphi(X-Y)]$. Thus, if $\varphi$ is rapidly decaying with distance (e.g. Gaussian kernel), the feature mappings
$\Phi(\mathbb{P})$ and $\Phi(\mathbb{P}')$ will be close to orthogonal (have a scalar product close to 0) whenever the distributions $\mathbb{P},\mathbb{P}'$ are well-separated. From this analysis, we anticipate that 
KMM with a Gaussian kernel will be robust against contaminations distributions $\Q_0$ whose main mass is far away from the source distributions, since its
representation $\Phi(\Q_0)$ will then
be close to orthogonal to $\bar{V}$ in feature space.

\section{Algorithm and applications}\label{sec:appli}
In this section, we will apply four methods on both synthetic and real datasets.\\
We choose to test three soft-DFM methods: KMM using the Energy Kernel \cite{kawakubo2016computationally}, our modified version of BBSE \cite{lipton2018detecting} and KMM using the Gaussian kernel \cite{iyer2014maximum} enhanced with Random Fourier Features, that we present in the next section. To show the benefit of the soft version, we also compare with one \textit{hard}-DFM method: KMM enhanced with Random Fourier Features.\\
The main objective of the experiments is, in view of our theoretical
results of Section~\ref{sub:mis} and
particularly Corollary~\ref{cor:main_contaminated}, to test robustness properties of several DFM methods against contamination of the the target dataset by different types of noise. Moreover, we want to check if the \textit{soft} version presented in Section \ref{sec:dfm} leads to improved results in some cases, and will not hurt the results in the others. \\
All the code and datasets are publicly available \cite{code}.
All the computations were done on a computer equipped with a NVIDIA RTX A2000 Laptop.

\subsection{Optimisation problem} \label{se:opt}

Whatever the chosen mapping, solving (\ref{P:DFM}) amounts to solving a Quadratic Programming (QP) in dimension $c$. Indeed, 
we can rewrite the problem as:
\begin{align*}
	\label{QP}
	\tag{QP}
	 & \text{minimise} \ \frac 1 2 \alpha^T \egram \alpha + q^T \alpha \\
	 & \text{subject to} \ \alpha \succeq 0_c \ \text{and} \ \textbf{1}_c^T \alpha = 1,
\end{align*}
with $q = \left(\langle\Phi(\hat{\mathbb{P}_i}), \Phi(\hat{\mathbb{Q}})\rangle\right)_{i=1}^c$. This is a $c$-dimensional QP problem, which can be solved efficiently.

The computational bottleneck is the computation of the Gram matrix $\egram$ and the vector $q$. Using KMM directly leads to a complexity of $O(n(n+m))$, as computing $q$ requires evaluating the kernel for every pair of points from the source and the target and computing $\egram$ requires evaluating the kernel for every pair of points between the source classes. Moreover, one needs to have permanent access to the source distributions, as computing $q$ requires both the source and target raw dataset.\\
Due to such issues, kernel matrix approximations are often used in order to reduce the computational cost of kernel methods \cite{camoriano2016nytro,rudi2015less}. In our case we use the well-established principle of Random Fourier Features (RFF) approximation \cite{rahimi2007random}.
RFF allows to obtain an approximation of a translation invariant kernel $k$, i.e.
$k(x,y) = \varphi(x-y)$, of the form:
$
\tilde{k}(x,y) = \tilde{\Phi}(x)^T\tilde{\Phi}(y),
\text{ with } \tilde{\Phi}(x) \in \mathbb{R}^D,
$
which is itself a positive semi-definite kernel. For a theoretical analysis of the uniform approximation quality of $k$ by $\tilde{k}$, see e.g. \cite{rudi2017generalization,sutherland2015error}. See the appendix \ref{appendix:rff} for more detail on Random Fourier Features.

Relying on RFF with $D$ Fourier features induces a complexity of $O(D(n+m))$ since we only have to compute $\tilde{\Phi}(\hat{\P}_i)$ and $\tilde{\Phi}(\hat{\Q})$. Computing $\tilde{\Phi}(\hat{\P})$ reduces to a matrix multiplication, for which GPU are well suited. To deal with memory overflow on GPU, we rely on the Python package \textit{PyKeops} \cite{JMLR:v22:20-275}.
With this implementation can solve (\ref{P:DFM}) for high-dimensional data with a very large number of points in sub-second times. For example, for two datasets containing $5\times10^6$ points in dimension $5$, (\ref{P:DFM}) is solved in less than a second, while almost 2 minutes are needed when we use the exact KMM.

In the experiments, we will design by Random Fourier Features Matching (RFFM) the DFM method that uses $\tilde{\Phi}$ as a feature mapping. RFFM can be used with any translation invariant kernel, but we choose to stick to the classical Gaussian kernel: $k(x,y) = \exp\left(-\|x-y\|^2/(2\sigma^2)\right)$ where the parameter $\sigma$ is optimised using the criterion derived from \eqref{theo:main:eq1}. Note that RFFM is only used as a way to speed up the computation, and hence we would obtain similar results with a classical KMM using the Gaussian kernel.

\subsection{Experiments} \label{experiments}

We want to test the robustness of the DFM methods in the contaminated scenario \ref{cond:CLS}. We will compare 4 methods: RFFM, softRFFM, softEnergy and softBBSE+. RFFM is the method introduced in the previous section while softRFFM is RFFM when we use the soft procedure introduced in Section \ref{sec:dfm}. SoftEnergy is the kernel mean matching method when we used the Energy kernel \cite{kawakubo2016computationally} and softBBSE+ is our version of BBSE \cite{lipton2018detecting}.\\
We will test the methods on both synthetic data and real datasets.


\subsection*{Gaussian Mixture}
In this setting, the source is a list of $c$ Gaussian distributions: $\mathbb{P}_1,\cdots,\mathbb{P}_c$. Our objective is to estimate $\alpha^*$ for different 
values of the contamination level $\epsilon = \alpha^*_0$ ranging from from $0$ to $0.3$ and different kinds of noise distributions $\mathbb{Q}_0$. We will test three kinds of noise (see Figure \ref{fig:all noises}):
\begin{enumerate}
    \item $\mathbb{Q}_0$ is uniformly distributed over the data range
    ("Background noise").
    \item $\mathbb{Q}_0$ is Gaussian with a mean distant from the other means.
    \item $\mathbb{Q}_0$ is Gaussian with a similar mean to the others. 
\end{enumerate}

\begin{figure*}[!ht]
\vskip 0.2in
\begin{center}
\centerline{\includegraphics[width=\textwidth]{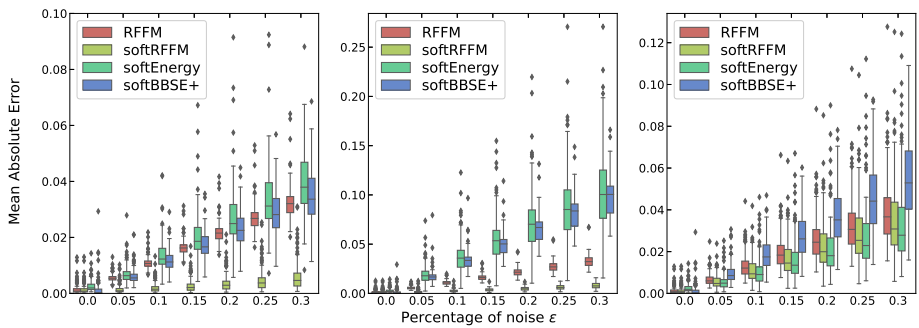}}
\caption{Robustness of the algorithms to three types of noise. (Lower is better.) Left: background noise; middle: noise is a new class far from the others; right: noise is a new class in the middle of the others. } 
\label{fig:robustness_results}
\end{center}
\vskip -0.2in
\end{figure*}

Throughout the experiments, we fix the number of classes in the source to $c=5$ and the number of non-contaminated points to $10000$. The dimension varies from $2$ to $10$. For each contamination level $\epsilon$ and each possible dimension,  we perform $20$ repetitions with different Gaussian distributions.
Results for the three experiments can be found in Figure \ref{fig:robustness_results}.

In the absence of noise contamination in the target, all methods give excellent results because the source distributions are easy to distinguish. Obviously, the results deteriorate as the contamination level increases. All methods still perform well against background noise (the loss is around $0.1$ in the worst case), with softRFFM being significantly better than the others. In the same fashion, when we add a Gaussian far away from the other distributions, softRFFM significantly outperforms the others. As discussed following Corollary~\ref{cor:main_contaminated}, this is because when
contamination puts mostly mass far from the other classes, the Gaussian KME of the noise distribution will be close to orthogonal to the source KMEs. This property does not hold when a class is added close to the others and thus can be more easily confounded. Thus, the results align well with our
theoretical analysis.

While Theorem \ref{theo:main_contaminated} and
Corollary~\ref{cor:main_contaminated} hold for the Energy kernel as well (assuming bounded data) or BBSE+, 
we don't observe robustness against noise.
Again, this is in line with the theoretical study for BBSE+ for which we expected no robustness. Concerning the Energy kernel, we surmise that the lack of robustness
comes from the fact that $k(x,y)$ can take 
large values even if $\|{x-y}\|$ is large, hence
near-orthogonality of the noise distribution to the
source does not hold in the corresponding feature space,
in contrast to the Gaussian kernel.


\subsection*{Cytometry dataset}
\begin{figure*}[!ht]
\vskip 0.2in
\begin{center}
\centerline{\includegraphics[width=0.8\textwidth]{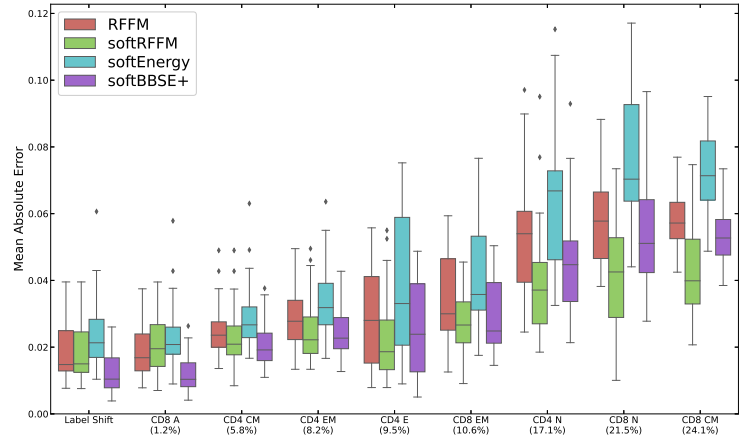}}
\caption{Each column represents the error — computed using the $\ell_2$ norm between the true proportions and the estimated proportions — obtained when some class is absent from the source but present in the target distribution. The first column gives the results when no class is discarded. The class are sorted according to the average proportions they represent in the samples (x labels mention class held out from the source and its proportion)}
\label{fig:hipc_results}
\end{center}
\vskip -0.2in
\end{figure*}

We test the robustness of our methods on the T-cell panel of the Human Immunology Project Consortium (HIPC) \cite{brusicComputationalResourcesHighdimensional2014,finak2016standardizing}. HIPC is composed of 63 samples. Seven laboratories analysed 3 replicates from 3 different patients. The number of measurements in the samples range from $10^4$ to $10^5$. The samples were manually separated into 10 categories using 7 markers.\\
To put ourselves in the \ref{cond:CLS} setting, we choose to remove one of the class of the source, so that this class becomes the noise in the target.
In detail, each of the patient replicates are joined into
a single patient sample, resulting in 3 patient joined samples for each of the seven laboratories. We take each sample successively as source, and patients samples
of the same laboratory as target.\footnote{There can be large variability between samples coming from different laboratories, while there is homogeneity within each
lab. The label shift hypothesis is therefore reasonable when keeping source and target from the same lab.} 
We first perform the experiment with all cell classes: in that case we can assume that we are in the vanilla Label Shift \ref{cond:LS} setting. Then, we repeat the experiment 10 times, removing each time a class from the source
but not from the target, thus playing the role of contaminant.
The results can be found in Figure \ref{fig:hipc_results}



We reach the same conclusion on the robustness of softRFFM compared to RFFM, softBBSE+ and softEnergy when we apply them on a noisy real-world dataset. This advantage of robustness is all the more significant as the proportion of noise is high.

\section{Conclusion}

We introduced Distribution Feature Matching (DFM) as a general approach
for class proportion estimation (also known at label shift estimation or
quantification learning), 
recovering 
methods from previous literature as special cases. We also
proposed the use of Random Fourier Features to speed up the computation
of kernel-based approaches and obtain an explicit finite-dimensional
vectorization (or "sketch") of the distributions.

We provided a general theoretical analysis of DFM, improving over previously 
known bounds derived for specific instantiations only. Furthermore, we analysed 
theoretically the behavior of DFM under departures from the
label shift hypothesis, a situation not studied in earlier works, and put into light a robustness against certain
types of perturbations, depending on the feature mapping used.
The theoretical analysis suggested a better robustness property of RFF
approaches based on a rapidly decaying translation-invariant kernel,
and this could be confirmed through numerical experiments on 
synthetic and real data.

Recent works \cite{tachet2020domain,zhang2013domain} 
considered a more general situation beyond label shift. In the {\em Generalized Label Shift} model of \cite{tachet2020domain}, the condition 
\eqref{cond:LS} does not hold in input space, but only
after transformation through
a suitable feature mapping $\Phi$, and these authors 
proposed an algorithm alternating between class proportion
estimation using BBSE+ and feature learning (using a separate domain adaptation
algorithm, suitably adapted to handle label proportion shift).
We believe that using one of the proposed methods in the present paper
could be used fruitfully in such a framework to replace the 
BBSE+ module.

\section*{Acknowledgements}
B. Dussap was supported by the program Paris Region Ph.D. of \href{https://www.dim-mathinnov.fr/}{DIM Mathinnov}. 
G. Blanchard acknowledges support of the ANR under ANR-19-CHIA-0021-01 ``BiSCottE'', IDEX REC-2019-044, and of the DFG under SFB1294 - 318763901.

\newpage

\section*{Ethical statement}

Label shift quantification has uses in a number of application domains; 
the results in this paper are chiefly oriented towards theory and 
general methodology so that we don't discuss an application in particular.
We only mention that the flow cytometry data used for proof-of-concept
experiments is publicly available from a reputable scientific consortium and has been up to our knowledge collected following all established ethical standards.

The original \textit{Classify \& Count} \cite{forman2005counting} method for label
shift quantification is known to inherit the potential biases of the
classification method it is based on (i.e. the misclassification errors can
be very unevenly distributed across classes and "favor" majority classes).
The \textit{Adjusted Classify \& Count (ACC)} approach and related methods \cite{forman2008quantifying,lipton2018detecting} aim at rectifying this bias.
In the present paper, we aim at going one step further and analyze certain robustness properties of the 
proposed label shift quantification methods, and introduce the contaminated
label shift \eqref{cond:CLS} setting with the goal of investigating
trustworthiness of such methods under mild violations of the standard
Label Shift model. Certainly the robustness property is desirable
for improved reliability in practice, but 
does not mean immunity against biases; additionally, the user should
always be wary of stronger model violations between reference and test
data, in particular class-conditional distribution shifts.
We therefore recommend established good practice of regularly checking on control
data possible biases or drifts from the model, in particular for sensitive 
applications.

\newpage
\bibliographystyle{splncs04}
\bibliography{article}

\newpage
\appendix

\section{Proofs}\label{appendix:prooftheo}

\subsection{Proof of Theorem \ref{theo:main}.}\label{proof:theo1}
To prove Theorem 1, we will require a result on the approximation error of the vector $\Phi(\mathbb{P})$ by an $n$-sample $\Phi(\hat{\mathbb{P}})$. For this, we use the 
vector Hoeffding's inequality (for a Euclidean or Hilbert norm),
a result which has appeared in different versions in the literature
(one of the earliest version seems to be from
\cite{pinelis1992approach}); see also e.g.
\cite{lopez2015towards} or \cite{smola2007hilbert}
in the KME setting.
While this result is not new, we give here a short self-contained proof for completeness.

\begin{theorem}\label{theo:approx}

Let $Z_1,\ldots,Z_n$ be independent (not necessarily identically
distributed) random variables taking values in a Hilbert space $\mathcal{H}$ (possibly $\mathcal{H} = \mathbb{R}^D$), such that $\forall i \in [n]: \|Z_i\| \leq C$ a.s.
    Then with probability greater than $1-\delta$ :
    \begin{equation*}
    \left\| \frac{1}{n} \sum_{i=1}^n (Z_i - \mathbb{E}[Z_i])\right\|
    \leq C\frac{(2 + \sqrt{2\log(1/\delta)})}{\sqrt{n}},
    \end{equation*}
\end{theorem}

\begin{proof}
 Since $\|Z_i\| \leq C$ we can assume without loss of generality
 that the random variables $Z_i$ take values in the ball $\mathcal{B}_C:= \{ z \in \mathcal{H}: \|z\| \leq C\}$.
 Define the function $F: (B_C)^n \rightarrow \mathbb{R}$ as
 \[
 F(z_1,\ldots,z_n) := \left\|\frac{1}{n} \sum_{i=1}^n (z_i -\mathbb{E}[Z_i]) \right\|.
 \]
    Straightforward computations show that the function $F$ satisfies the bounded difference condition. Namely, let us fix all the values $z_1, \cdots, z_n$ in $\mathcal{B}_C$ except for the $z_j$ which will be set to $\Bar{z}_j$; then
    \begin{align*}
        \left| F(z_1, \cdots, z_n) - F(z_1, \cdots, \Bar{z}_j, \cdots ,z_n) \right| 
        = \frac 1 n \left\|z_i - \Bar{z}_i\right\| 
        \leq \frac{2C}{n}.
    \end{align*}
    Using McDiarmid's inequality,
    since $Z_i$ are realisations 
    of independent random variables taking values in $\mathcal{B}_C$,
    it holds with probability greater than $1-\delta$:
    \begin{equation*}
    \left\| \frac{1}{n} \sum_{i=1}^n (Z_i - \mathbb{E}[Z_i])\right\|
    \leq \mathrm{E} \left[ \left\| \frac{1}{n} \sum_{i=1}^n (Z_i - \mathbb{E}[Z_i])\right\| \right]
    + C\sqrt{\frac{2\log(1/\delta)}{n}}.
    \end{equation*}
   Let us write $\tilde{Z}_i = Z_i - \mathbb{E}[Z_i]$; $\tilde{Z}_i$ are also independent random variables,
   bounded in norm by $2C$, and are centered.
    We only have to bound the 
    expectation in the right-hand side above.

    By Jensen's inequality,
    \begin{align*}
        \mathbb{E}\left[\left\| \frac 1 n \sum \tilde{Z}_i \right\|\right] &\leq
        \sqrt{\mathbb{E}\left[\left\| \frac 1 n \sum \tilde{Z}_i \right\|^2\right]}
        \\
        &= \frac{1}{n} \left( 
        \sum_{i,j=1}^n \mathbb{E}[ 
        \langle \tilde{Z}_i,\tilde{Z}_j \rangle ]
        \right)^{\frac{1}{2}}
        \\
        &= \frac{1}{n}\left( 
        \sum_{i=1}^n \mathbb{E}[ 
        \| \tilde{Z}_i \|^2]
        \right)^{\frac{1}{2}}
        \leq \frac{2C}{\sqrt{n}},
    \end{align*}

where we have used that since for $i\neq j$ the variables
    $\tilde{Z}_i,\tilde{Z}_j$ are independent and centered it holds
    $\mathbb{E}\left[\langle \tilde{Z}_i, \tilde{Z}_j\rangle\right] = 0$.
    \qed
\end{proof}

We now turn to the proof of Theorem \ref{theo:main}.

\begin{proof}
    \label{proof:theo:main}
    Let $\alpha \in \Delta^c$ be fixed. Later we will
    consider the choices $\alpha = \alpha^*$ or $\alpha = \tilde{\alpha}$ (the population resp. empirical class proportions in the target domain), and will specify this when needed, but a large part of the argument holds for
    any $\alpha$.
    
    Given a feature map $\Phi$, let us use the notation introduce in definition \ref{def:gram}.
    Let us note $D_\Phi$ the function defined by $D_\Phi(\P_1, \P_2) = \|\Phi(\P_1) - \Phi(\P_2) \|$. Note that $D_\Phi$ will be a distance if and only if the mapping $\Phi$ is injective and will a pseudo-distance otherwise.\\
    
    It holds
    \begin{align*}
    	    D_\Phi\left(\displaystyle\sum_{i=1}^c \hat{\alpha}_i \hat{\mathbb{P}}_i, \displaystyle\sum_{i=1}^c \alpha_i \hat{\mathbb{P}}_i\right)^2 
            &= \left\| \displaystyle\sum_{i=1}^c \hat{\alpha}_i \Phi(\hat{\mathbb{P}}_i) - \displaystyle\sum_{i=1}^c \alpha_i \Phi(\hat{\mathbb{P}}_i) \right\|^2 \\
    	    &= \left\| \displaystyle\sum_{i=1}^c (\hat{\alpha}_i - \alpha_i) \Phi(\hat{\mathbb{P}}_i) \right\|^2 \\
    	    &= (\hat{\alpha} - \alpha) \ \egram \ (\hat{\alpha} - \alpha) \\
    	    &\geq \left(\minortho \ u^T \egram u \right) \ \| \hat{\alpha} - \alpha \|^2 \\
            &\stackrel{(\dagger)}{=} \lambdamin(\egram) \ \| \hat{\alpha} - \alpha \|^2,
    \end{align*}
    with equality $(\dagger)$ proven in Theorem  \ref{theo appendixC delta}.
		
	Thus in order to bound $\|\hat{\alpha} - \alpha \|$ we have to upper-bound: $D_\Phi(\sum_{i=1}^c \hat{\alpha}_i \hat{\mathbb{P}}_i, \sum_{i=1}^c \alpha_i \hat{\mathbb{P}}_i)$.
	By the triangle inequality, this is upper-bounded by $D_\Phi(\sum_{i=1}^c \hat{\alpha}_i \hat{\mathbb{P}}_i, \hat{\mathbb{Q}}) + D_\Phi(\hat{\mathbb{Q}}, \sum_{i=1}^c \alpha_i \hat{\mathbb{P}}_i)$.
	By definition of $\hat{\alpha}$, we have $D_\Phi(\sum_{i=1}^c \hat{\alpha}_i \hat{\mathbb{P}}_i, \hat{\mathbb{Q}}) \leq D_\Phi(\sum_{i=1}^c \alpha_i \hat{\mathbb{P}}_i, \hat{\mathbb{Q}})$. Hence, we can upper-bound the quantity by $2 D_\Phi(\sum_{i=1}^c \alpha_i \hat{\mathbb{P}}_i, \hat{\mathbb{Q}})$. 

	Using the triangle inequality once again, we can upper bound the previous quantity by $D_\Phi\left(\sum_{i=1}^c \alpha_i \hat{\mathbb{P}}_i, \sum_{i=1}^c \alpha_i \mathbb{P}_i\right) + D_\Phi\left(\sum_{i=1}^c \alpha_i \mathbb{P}_i, \hat{\mathbb{Q}} \right)$.

	Using Theorem \ref{theo:approx} and the union bound, it holds with probability greater than  $1-\delta/2$:
 
     \begin{align} \label{eq:term1} 
     D_\Phi\left(\displaystyle\sum_{i=1}^c \alpha_i \hat{\mathbb{P}}_i, \displaystyle\sum_{i=1}^c \alpha_i \mathbb{P}_i\right) 
	    &\leq \displaystyle\sum_{i=1}^c \alpha_i D_\Phi(\hat{\mathbb{P}}_i, \mathbb{P}_i) \nonumber \\
	    &\leq \sum_{i=1}^c \alpha_i C \frac{2 + \sqrt{2\log{2c/\delta}}}{\sqrt{n_i}}\nonumber \\
        &= CR_{c/\delta} \sum_{i=1}^c \frac{\alpha_i}{\sqrt{n_i}},
     \end{align}
    with $R_x = 2 + \sqrt{2\log{2x}}$.\\
    Since $n_i = n\beta_i$, it holds $\sum_{i=1}^c  \frac{\alpha_i}{\sqrt{n_i}} = \frac{1}{\sqrt{n}} \sum_{i=1}^c \frac{\alpha_i}{\sqrt{\tilde{\beta_i}}}$. If we write $w_i = \frac{\alpha_i}{\tilde{\beta_i}}$, then by Hölder's inequality: 
	\begin{equation} \label{eq:term1hol}
	\sum_{i=1}^c \frac{\alpha_i}{\sqrt{\tilde{\beta_i}}} = \sum_{i=1}^c \left(w_i \sqrt{\tilde{\beta_i}}\right) \leq \|w\|_2 \underbrace{\sqrt{\sum_{i=1}^c \tilde{\beta_i}}}_{=1} .
	\end{equation}

    We finally turn to bounding the term 
	$D_\Phi\left(\sum_{i=1}^c \alpha_i \mathbb{P}_i, \hat{\mathbb{Q}} \right)$. This is the only point of the proof where we need to use the label shift assumption  and specify $\alpha$. First, under ($\mathcal{LS}$)
    this is equal to 
    $D_\Phi\left(\sum_{i=1}^c \alpha_i \mathbb{Q}_i, \hat{\mathbb{Q}} \right)$. We now distinguish between two possibilities:
    \begin{itemize}
    \item If $\alpha=\alpha^*$, then $\sum_{i=1}^c \alpha^*_i \mathbb{Q}_i = \mathbb{Q}$, so
    that using Theorem \ref{theo:approx}, it holds with probability greater than  $1-\delta/2$:    
    \begin{equation} \label{eq:term2a}
    D_\Phi\left(\sum_{i=1}^c \alpha_i^* \mathbb{Q}_i, \hat{\mathbb{Q}} \right)
    = D_\Phi\left(\mathbb{Q}, \hat{\mathbb{Q}} \right) \leq C\frac{R_{1/\delta}}{\sqrt{m}}.
    \end{equation}
    \item For $\alpha=\tilde{\alpha}$, 
    it holds $\tilde{\alpha}_i = m_i/m$,
    where $m_i$ is the number of target sample points of class $i$. We then get
    \begin{align*}
    D_\Phi\left(\sum_{i=1}^c \tilde{\alpha}_i \mathbb{Q}_i, \hat{\mathbb{Q}} \right)
    & = \left \| \frac{1}{m}\sum_{i=1}^c m_i \Phi( \mathbb{Q}_i) - \frac{1}{m} \sum_{j=n+1}^{n+m} \Phi(X_j) \right\| \\
    & = \left \| \frac{1}{m} \sum_{j=n+1}^{n+m} \Phi(\mathbb{Q}_{Y_j}) - \frac{1}{m} \sum_{j=n+1}^{n+m} \Phi(X_j) \right\| \\
    &=
    \left\| \frac{1}{m}\sum_{j=n+1}^{n+m} ( \Phi(X_j) - \Phi(\mathbb{Q}_{Y_j}) )\right\|.
    \end{align*}
    Now, notice that conditionally to the labels $(Y_j)_{j=n+1}^{n+m}$, the target sample points
    $X_j$ are independent, not identically
    distributed but with respective class
    conditional distribution
    $\mathbb{Q}_{Y_j}$. We can therefore still
    appeal to Theorem \ref{theo:approx}, and
    conclude that it holds with probability greater than  $1-\delta/2$:
    \begin{equation} \label{eq:term2b}
    \left\| \frac{1}{m}\sum_{j=n+1}^{n+m} ( \Phi(X_j) - \Phi(\mathbb{Q}_{Y_j}) )\right\| 
    \leq C\frac{R_{1/\delta}}{\sqrt{m}}.
    \end{equation}
    \end{itemize}
    In both cases we therefore get the same bound for this last term.

    Putting together \eqref{eq:term1}, \eqref{eq:term1hol} and either \eqref{eq:term2a} (for $\alpha=\alpha^*$) or \eqref{eq:term2b} (for $\alpha=\tilde{\alpha}$) and $R_{1/\delta} \leq R_{c/\delta}$, gives the first claimed inequality of the theorem.\\
 
	To obtain the second claimed inequality, we can see that the worst case scenario for $w$ is obtain when $\alpha_i=1$ for the smallest $\tilde{\beta_i}$ and $0$ for the others. Hence, $\frac{1}{\sqrt{n}} \|w\|_2 \leq \underset{i}{\text{max}}  \ \frac{1}{\sqrt{n_i}}$.\qed
	\end{proof}

\begin{remark}
    The best case scenario for $\|w\|_2$ is obtain when either $\alpha^*$ (or $\tilde{\alpha}$) equals $\tilde{\beta}$ or when $\beta_i = \frac 1 n$.
    In both cases, $\|w\|_2 = \sqrt{c}$.
\end{remark}

\begin{corollary}
    The result of Theorem 1 still holds, with $\deltamin$ replace by $\lambdamin$, if we add a "dummy" class whose mapping is equals to $0$, in other words, if we solve the soft-DFM version.
\end{corollary}

\begin{proof}
    Using the same notation introduced in the proof of Theorem \ref{theo:main}, we have:
    \begin{align*}
        D_\Phi^2\left(\displaystyle\sum_{i=0}^c \hat{\alpha}_i \hat{\mathbb{P}}_i, \displaystyle\sum_{i=0}^c \alpha_i \hat{\mathbb{P}}_i\right) &= (\hat{\alpha} - \alpha) \ \egram \ (\hat{\alpha} - \alpha) \\
        &\geq \left(\underset{\|u\|=1}{\text{min}} \ u^T \egram u \right) \ \| \hat{\alpha} - \alpha \|^2 \\
        &= \lambdamin \ \| \hat{\alpha} - \alpha \|^2
    \end{align*}
    with $\alpha_0 := 0$ and $\underset{\|u\|=1}{\text{min}} \ u^T \egram u $ the smallest eigenvalue of the Gram matrix $\egram$.
    

    From this point on, we can follow the rest of the proof of Theorem 1.
    \qed
\end{proof}

\subsection{Theorem \ref{theo:main_contaminated}}\label{proof:theo2}

Let us first prove a lemma.

\begin{lemma}\label{lemma:proj}
    Let $\mathcal{H}$ be a Hilbert space, $\cC$ be a closed convex subset and $V$ an affine subspace of $\cH$ such that $\cC \subset V \subset \mathcal{H}$. For every $x\in\mathcal{H}$ we have 
    \[
    \Pi_\cC(x) = \Pi_\cC(\Pi_V(x)),
    \]
    with $\Pi_\cC$ and $\Pi_V$ the minimum distance projection functions onto $\cC$ and $V$.
\end{lemma}

\begin{proof}
    Let us take $x \in \mathcal{H}$ and $c \in \cC$. Note $p = \Pi_V(x)$, since $c \in V$ we can use Pythagoras' theorem : 
    \[
    	\|x-c\|^2 = \|x-p\|^2 + \|p-c\|^2.
    \] 
    The point $c$ that minimises $\|x-c\|^2$, i.e $\Pi_\cC(x)$, is the same point $c$ that minimises $\|p-c\|^2$, i.e $\Pi_\cC(\Pi_V(x))$.
    \qed
\end{proof}

Let us now prove Theorem \ref{theo:main_contaminated}.

\begin{proof}
        Below, we use the notation $\alpha$ for $\alpha^*$.\\
		Let us use the notation $D_{\Phi}(\P_1, \P_2) = \|\Phi(\P_1) - \Phi(\P_2) \|$.\\ 
		We write $\cC_n = \mathrm{ConvHull}\langle \Phi(\hat{\mathbb{P}}_1), \cdots, \Phi(\hat{\mathbb{P}}_c) \rangle$, the convex hull of the mapped empirical distributions $\hat{\mathbb{P}}_i$, and $\Pi_{\cC_n}$ the projection onto this convex. In the same fashion let us write $\cC = \mathrm{ConvHull}\langle \Phi(\mathbb{P}_1), \cdots, \Phi(\mathbb{P}_c) \rangle$ and the associated projection $\Pi_\cC$, and finally $V$ the affine subspace generated by the $\Phi(\mathbb{P}_i)$, namely
        $V = \{ \sum_{i=1}^c \lambda_i \Phi(\mathbb{P}_i) \ | \ \sum_{i=1}^c \lambda_i = 1\}$.
        We will sometimes write $\Pi_\cC(\mathbb{Q})$ as a shorthand for $\Pi_\cC(\Phi(\mathbb{Q}))$.
		
        With the same computation as in the proof of Theorem \ref{theo:main}, we have:
        \[
            \| \hat{\alpha} - \alpha \| \leq \frac{1}{\sqrt{\deltamin}} \ D_{\Phi}\left(\displaystyle\sum_{i=1}^c \hat{\alpha}_i \hat{\mathbb{P}}_i , \displaystyle\sum_{i=1}^c \alpha_i \hat{\mathbb{P}}_i\right).
        \]
        Once again we have to upper-bound $D_{\Phi}\left(\sum_{i=1}^c \hat{\alpha}_i \hat{\mathbb{P}}_i , \sum_{i=1}^c \alpha_i \hat{\mathbb{P}}_i\right)$. By the triangle inequality:
        \begin{small}
        \begin{equation}\label{eq:pr1.4}
        D_{\Phi}\left(\sum_{i=1}^c \hat{\alpha}_i \hat{\mathbb{P}}_i , \sum_{i=1}^c \alpha_i \hat{\mathbb{P}}_i\right) 
        \leq
        \underbrace{
            \left\|\sum_{i=1}^c \hat{\alpha}_i \Phi(\hat{\mathbb{P}}_i) - \Pi_\cC(\mathbb{Q})\right\|
        }_{\text{(1)}}
        + \underbrace{
            \left\|\Pi_\cC(\mathbb{Q}) - \sum_{i=1}^c \alpha_i \Phi(\hat{\mathbb{P}}_i)\right\|
        }_{\text{(2)}}.
        \end{equation}
        \end{small}
        Let us analyse the second term first.
        We use the triangle inequality:
        $$
        \left\|\Pi_\cC(\mathbb{Q}) - \sum_{i=1}^c \alpha_i \Phi(\hat{\mathbb{P}}_i)\right\| \leq \left\|\Pi_\cC(\mathbb{Q}) - \sum_{i=1}^c \alpha_i \Phi(\mathbb{P}_i)\right\| + D_{\Phi}\left(\sum_{i=1}^c \alpha_i \mathbb{P}_i , \sum_{i=1}^c \alpha_i \hat{\mathbb{P}}_i\right).
        $$
        Using \ref{eq:term1} and \ref{eq:term1hol}, the second term can be bounded by 
        $CR_{c/\delta} \frac{\|w\|_2}{\sqrt{n}}$, with probability greater than $1-\delta/2$,
        as we did in the proof of Theorem \ref{theo:main}.
        For the first term we will require three elements:
        
        \begin{enumerate}
            \item $\sum_{i=1}^c \alpha_i \Phi(\mathbb{P}_i)$ lies in $C$.
            \item For all $x$, $\Pi_\cC(x) = \Pi_\cC(\Pi_V(x))$, see Lemma \ref{lemma:proj}.
            \item $\Pi_\cC$ is a contraction.
        \end{enumerate}
        
        With that in mind, 
        \begin{align*}
            \left\| \sum_{i=1}^c \alpha_i \Phi(\mathbb{P}_i) - \Pi_\cC\left(\Phi(\mathbb{Q})\right) \right\| &=
            \left\| \Pi_\cC\left(\sum_{i=1}^c \alpha_i \Phi(\mathbb{P}_i)\right) - \Pi_\cC( \Pi_V(\Phi(\mathbb{Q})))\right\| \\
            &\leq \left\| \sum_{i=1}^c \alpha_i \left( \Phi(\mathbb{P}_i)  - \Pi_V(\Phi(\mathbb{Q}_i) \right) \right\| \\
            &\leq \max_i \left\| \Phi(\mathbb{P}_i)  - \Pi_V(\Phi(\mathbb{Q}_i))  \right\|.
        \end{align*} 
        Hence, we have 
        \[
        \left\|\Pi_\cC(\mathbb{Q}) - \sum_{i=1}^c \alpha_i \Phi(\hat{\mathbb{P}}_i)\right\| 
        \leq CR_{c/\delta} \frac{\|w\|_2}{\sqrt{n}} 
        + \underbrace{
            \max_i\ \| \Phi(\mathbb{P}_i)- \Pi_V(\Phi(\mathbb{Q}_i))\|
            }_{B^\parallel(\mathbb{P}, \mathbb{Q})}.
        \]
        Let us turn to the first term of \eqref{eq:pr1.4}. By definition of $\hat{\alpha}$, it holds $\sum_{i=1}^c \hat{\alpha}_i \Phi(\hat{\mathbb{P}}_i) = \Pi_{\cC_n}(\Phi(\hat{\mathbb{Q}}))$.
        Using the triangle inequality
        \[
        \left\|\Pi_{\cC_n}(\hat{\mathbb{Q}}) - \Pi_{\cC}(\mathbb{Q}))\right\| \leq 
        \left\|\Pi_{\cC_n}(\hat{\mathbb{Q}}) -\Pi_{\cC_n}(\mathbb{Q})\right\|
        + \left\Vert\verticalspace\Pi_{\cC_n}(\mathbb{Q}) - \Pi_\cC(\mathbb{Q})\right\Vert.
        \]
        Since $\Pi_{\cC_n}$ is a contraction, we have $\left\|\Pi_{\cC_n}(\hat{\mathbb{Q}}) - \Pi_{\cC_n}(\mathbb{Q})\right\| \leq D_{\Phi}(\mathbb{Q}, \hat{\mathbb{Q}}).$\\
        
        The only thing left to bound is the term $\| \Pi_{\cC_n}(\Phi(\mathbb{Q})) - \Pi_\cC(\Phi(\mathbb{Q})) \|$. For this we use
        results of \cite{alber1993some} relating the
        distance between convex projections onto two
        different convex sets in relation to
        their Hausdorff distance. Recall the definition of the Hausdorff distance between two sets:

        \begin{definition}
            Let $X$ and $Y$ be two non-empty subsets of a metric space $(M, d)$. We define their Hausdorff distance $H(X, Y)$ by:
            \[ 
            H(X, Y)= \max \left\{\,\sup _{x\in X}d(x,Y),\,\sup _{y\in Y}d(X,y)\,\right\}
            \]
        \end{definition}
        
        Using Theorem 3.6 and Remark 3.7 of \cite{alber1993some}, the latter
        quantity is smaller than $\sqrt{2 H(\cC, \cC_n)(r+d)}$, where $r = \text{dist}(0,\Phi(\mathbb{Q}))$, $d = \text{max}\{ \text{dist}(0, \cC),  \text{dist}(0, \cC_n)\}$ and $0$ the origin. 
      	Since the problem has a geometrical nature and is invariant by translation, we can translate everything
        so that $\Phi(\mathbb{Q})$ is the origin of the space. Hence, the bound reads $\sqrt{2 H(\cC, \cC_n)d}$ with $d = \text{max}\{ \text{dist}(\Phi(\mathbb{Q}), \cC),  \text{dist}(\Phi(\mathbb{Q)}, \cC_n)\}$. Let us take care of the Hausdorff distance first:
        \begin{align*}
            \sup_{x \in \cC} \ d(x, \cC_n) &=  \sup_{x \in \cC} \| x - \Pi_{\cC_n}(x) \| \\ 
            &= \sup_{\lambda \in \Delta^c} \inf_{\beta \in \Delta^c} \left\|\sum_i \lambda_i \Phi(\mathbb{P}_i) - \sum_i \beta_i \Phi(\hat{\mathbb{P}}_i) \right\| \\
            &\leq \sup_{\lambda \in \Delta^c} \left\|\sum_i \lambda_i \Phi(\mathbb{P}_i) - \sum_i \lambda_i \Phi(\hat{\mathbb{P}}_i) \right\| \\
            &\leq \max_i \ D_{\Phi}(\mathbb{P}_i, \hat{\mathbb{P}}_i).
        \end{align*}
        A similar argument holds for $\sup_{x \in \cC_n} \ d(x, \cC)$, and hence $H(\cC, \cC_n) \leq \max_i D_{\Phi}(\mathbb{P}_i, \hat{\mathbb{P}}_i)$.\\
        
        We could simply bound $d$ by $2$ but we would obtain a loose bound. Instead,
        if we write $\Pi_\cC(\Phi(\mathbb{Q})) = \sum_{i=1}^c \lambda_i \Phi(\mathbb{P}_i)$ then \begin{align*}
            d(\Phi(\mathbb{Q}), \cC_n) &= \|\Phi(\mathbb{Q}) - \Pi_{\cC_n}(\Phi(\mathbb{Q}))\| \\
            &\leq \|\Phi(\mathbb{Q}) - \sum_{i=1}^c \lambda_i \Phi(\hat{\mathbb{P}}_i)\| \\
            &\leq \|\Phi(\mathbb{Q}) - \Pi_\cC(\Phi(\mathbb{Q}))\| + \left\| \sum_{i=1}^c \lambda_i \Phi(\mathbb{P}_i) - \sum_{i=1}^c \lambda_i \Phi(\hat{\mathbb{P}}_i)\ \right\| \\
            &\leq \|\Phi(\mathbb{Q}) - \Pi_\cC(\Phi(\mathbb{Q}))\| + \underset{i}{\text{max}} \ D_{\Phi}(\mathbb{P}_i, \hat{\mathbb{P}}_i).
        \end{align*}        
        Finally, using \ref{theo:approx}, we get with probability higher than $1-\delta/2$:
        \begin{align*}
            \sqrt{2H(\cC, \cC_n)d} 
            &\leq \sqrt{2} \ \max_i\ D_{\Phi}(\mathbb{P}_i, \hat{\mathbb{P}}_i) + \sqrt{2\max_i D_{\Phi}(\mathbb{P}_i, \hat{\mathbb{P}}_i) \ \|\Phi(\mathbb{Q}) - \Pi_\cC(\Phi(\mathbb{Q}))\|} \\
            &= \sqrt{2} \ \max_i\ D_{\Phi}(\mathbb{P}_i, \hat{\mathbb{P}}_i) + 
            \sqrt{2\max_i D_{\Phi}(\mathbb{P}_i, \hat{\mathbb{P}}_i)} \ B^\parallel(\mathbb{P}, \mathbb{Q}). \\
            &\leq \sqrt{2} \ \max_i\ \frac{CR_{c/\delta}}{\sqrt{n_i}} + \sqrt{2 \max_i\ \frac{CR_{c/\delta}}{\sqrt{n_i}}} \ B^\perp(\mathbb{P}, \mathbb{Q}). 
        \end{align*}    
		By putting everything together, with probability at least $(1-\delta)$, we have:
		\begin{align*}
		    \| \hat{\alpha} - \alpha^* \| &\leq \frac{1}{\sqrt{\deltamin}} 
        \Big(
            \frac{CR_{1/\delta}}{\sqrt{m}} + \sqrt{2} \ \max_i\ \frac{CR_{c/\delta}}{\sqrt{n_i}} + \sqrt{2 \max_i\ \frac{CR_{c/\delta}}{\sqrt{n_i}}} \ B^\perp(\mathbb{P}, \mathbb{Q}) 
            \\
            & \hspace{2.1cm} + CR_{c/\delta} \frac{\|w\|_2}{\sqrt{n}} + B^\parallel(\mathbb{P}, \mathbb{Q})
            ) 
        \Big) \\
        &\leq \frac{1}{\sqrt{\deltamin}} 
        \Big(
            c_1 \ \max_i\ \frac{CR_{c/\delta}}{\sqrt{n_i}} + \frac{CR_{1/\delta}}{\sqrt{m}} + \sqrt{2 \max_i\ \frac{CR_{c/\delta}}{\sqrt{n_i}}} \ B^\perp + B^\parallel
        \Big)
        ,
		\end{align*}
        with $c_1 = 1 + \sqrt{2}\leq 3.$
        \qed
\end{proof}

\subsection{Corollary \ref{cor:main_contaminated}}\label{proof:cor}

We directly apply Theorem 2 with the "dummy" class $\phi(\mathbb{P}_0):=0$. In that case, if we write $\bgram$ the Gram matrix of $\{\phi(\mathbb{P}_0), \phi(\hat{\mathbb{P}}_1) \cdots \phi(\hat{\mathbb{P}}_c)\}$ then, as the first column of this matrix is zero,
\[
\deltamin(\bgram) = \minortho\ u^T\bgram u = \underset{\|u\|=1}{\mathrm{min}} \ u^T\egram u = \lambdamin.
\]
In the same fashion, the affine subspace $V:=\mathrm{AffSpan}\{ \Phi(\mathbb{P}_i), i \in [0, \cdots, c]\}$ is equal to $\bar{V}:=\mathrm{Span}\{\Phi(\mathbb{P}_i), i \in [c]\}$. Finally, the convex hull $\mathrm{ConvHull}\{ \Phi(\mathbb{P}_i), i \in [0, \cdots, c]\}$ is equal to $\mathrm{int}(C)$: the interior of the convex hull $\mathrm{ConvHull}\{ \Phi(\mathbb{P}_i), i \in [c]\}$ \\
As we are in the contaminated label shift setting, $\mathbb{P}_i = \mathbb{Q}_i$ and hence 
\[
B^\parallel(\mathbb{P}, \mathbb{Q}) = 
\max_i \|\Phi(\mathbb{P}_i) - \Pi_{\bar{V}}( \Phi( \mathbb{Q}_i)) \|_{\mathcal{F}} 
    = \| \Pi_{\bar{V}}( \Phi( \mathbb{Q}_0)) \|
\]
The "orthogonal" term $B^\perp(\mathbb{P}, \mathbb{Q})$ can be bounded by $\sqrt{\alpha_0 \ \|\Phi( \mathbb{Q}_0 ) \|}$ as follows:
\begin{align*}
    B^\perp(\mathbb{P}, \mathbb{Q})^2 &= \| \Phi(\mathbb{Q}) - \Pi_{\mathrm{int}(C)}(\Phi(\mathbb{Q})) \| \\
    &\overset{(\dagger)}{\leq} \| \sum_{i=1}^c \alpha_i^*\phi(\mathbb{P}_i) + \alpha_0 \phi(\mathbb{Q}_0)  - \sum_{i=1}^c \alpha_i^* \phi(\mathbb{P}_i) \| \\
    &= \alpha_0 \| \phi(\mathbb{Q}_0) \|,
\end{align*}
for the inequality $(\dagger)$, we use the fact that $\sum_{i=1}^c \alpha_i^* \phi(\mathbb{P}_i) \in \mathrm{int}(C)$.

\begin{remark}
    Both theorem \ref{theo:main_contaminated} and corollary \ref{cor:main_contaminated} are true if we replace $\alpha^*$ by $\tilde{\alpha}$. 
\end{remark}

\subsection{Proposition \ref{prop:bbse}.}\label{proof:prop:bbse}

\begin{proof}
It is straightforward to check that 
for the mentioned feature mapping,
$\Phi(\hat{\mathbb{P}}_i)$ in the DFM setting
is exactly the $i$-th column of $M$ in the BBSE notation, and $\Phi(\hat{\mathbb{Q}})=Y$.
Hence the DFM objective in ($\mathcal{P}$) rewrites to $\| \alpha^T M - Y\|^2$, and since
$M$ is assumed invertible, in that setting the unconstrained
solution is $\hat{\alpha} = M^{-1}Y$.
Furthermore, the sum-1 condition $\bm{1}^T \alpha=1$ (where $\bm{1}$ denotes a vector of ones of dimension $c$) is automatically satisfied for the unconstrained solution: obviously it holds 
$\bm{1}^T M = \bm{1}^T$, hence $\bm{1}^T =\bm{1}^T M^{-1}$,
and $\bm{1}^T Y =1$, so that
$\bm{1}^T M^{-1} Y = 1$.
\end{proof}

\newpage

\section{Properties of $\deltamin$}\label{appendix:Delta}

Let $(b_1, \cdots, b_c)$  be a $c$-tuple of vectors of $\mathbb{R}^D$ assumed to be linearly independent and $\Bar{b}$ their mean. We denote $M$ the Gram matrix of those vectors, i.e. $M_{ij} = \langle b_i, b_j \rangle$. We also write $M^c$ the centered Gram matrix of the vectors : $(M^c)_{i,j} = \langle b_i - \Bar{b}, b_j - \Bar{b} \rangle$. Finally, we denote $\bm{1}$ a vector of ones (of dimension $c$).

In this appendix we prove two claims about the quantity $\deltamin := \deltamin(b_1, \cdots, b_c)$ defined in definition \ref{def:gram} as the second smallest eigenvalue of $M^c$.

\begin{theorem}
    \label{theo appendixC delta}
    For any number of classes $c$, $\deltamin$ is equal to $\minortho u^T M u$.
\end{theorem}

\begin{proof}
    Let us take $u$ such that $\|u\| = 1$, $\bm{1}^Tu = 0$ and $P$ the projection matrix on $<\bm{1}>^\perp$, such that $Pu = u$. We have:
    \begin{align*}
        u^TMu &= (Pu)^TM(Pu) \\ 
              &= u^T(PMP)u.
    \end{align*}

    Observe that $PMP$ is a symmetric matrix of rank $c-1$, hence the eigenvectors $v_i$ (associated to the eigenvalues $\lambda_1 \geq \lambda_2 \cdots \geq \lambda_c = 0$) form an orthonormal base.
    In particular $v_c = c^{-1/2}\bm{1}$. Since $u \in <\bm{1}>^\perp$, then $u \in <v_1, \cdots, v_{c-1}>$. There exist $\alpha \in \mathbb{R}^{c-1}$ such that $u = \sum_{i=1}^{c-1} \alpha_i v_i$, and since $\|u\|=1$ then $\|\alpha\|=1$. \\
    With that in mind :
    \begin{align*}
        u^T(PMP)u &= \left(\sum_{i=1}^{c-1} \alpha_i v_i\right)^T\left(\sum_{i=1}^{c-1} \alpha_i (PMP)v_i\right) \\
                  &= \left(\sum_{i=1}^{c-1} \alpha_i v_i\right)^T\left(\sum_{i=1}^{c-1} \alpha_i \lambda_i v_i\right) \\
                  &= \sum_{i,j=1}^{c-1} \lambda_i \alpha_i\alpha_j \langle v_i, v_j \rangle \\
                  &= \sum_{i=1}^{c-1} \lambda_i \alpha_i^2.
    \end{align*}
    Hence $\Delta_{\text{min}}(M)$ is equal to $\underset{\|\alpha\|=1}{\text{min}} \sum_{i=1}^{c-1} \lambda_i \alpha_i^2$. Using the change of variable $\beta_i = \alpha_i^2$, this is equivalent to find :$\underset{\beta \in \Delta^{c-1}}{\text{min}} <\lambda, \beta>$, which is equal to $\lambda_{c-1}$.
    All that is left to do is a straightforward computation of $PMP$ with $P = I_c - \frac{1}{c} \ \bm{1}^T\bm{1}$, to find that $(PMP)_{i,j} = \langle b_i - \Bar{b}, b_j - \Bar{b}\rangle$.
\end{proof}

A direct corollary of this theorem is that $\lambdamin$ is greater than the smallest singular value of the matrix $\left(b_1, \cdots, b_c \right)$.

\begin{theorem}
    \label{appendixC:lambdamin2classes}
	In particular for two classes, $\lambdamin(b_1, b_2) = \frac 1 2 \|b_1 - b_2\|^2$.
\end{theorem}

\begin{proof}
	In two dimensions, the conditions $\|x\|=1$ and $\bm{1}^Tx = 0$ can only be verified for 2 points, $x = \left(\sqrt{\frac 1 2}, - \sqrt{\frac 1 2}\right)$ and $x = \left(-\sqrt{\frac 1 2}, \sqrt{\frac 1 2}\right)$. If we compute $x^T M x$ for these two points we obtain : $\frac 1 2 \|b_1\|^2 - \langle b_1, b_2 \rangle  + \frac 1 2 \|b_1\|^2$ which is equal to $\frac 1 2 \|b_1 - b_2\|^2$.
\end{proof}

\newpage

\section{Short review of Random Fourier Features}\label{appendix:rff}
Random Fourier Features are based on Bochner's Theorem:
\begin{theorem}
    \label{theo:boch}
    A continuous function $\varphi$ on $\mathbb{R}^D$ is positive definite if and only if $\varphi$ is the Fourier transform of a non-negative measure.
\end{theorem}

A direct corollary of this result is that every continuous invariant kernel $k$ (associated to a function $\varphi$) is the Fourier transform of a non-negative measure that we denote $\Lambda_k$. One can show that $\Lambda_k(\mathbb{R}^D) = \varphi(0)$. Hence, for a normalized continuous invariant kernel, $\Lambda_k$ is a distribution referred to as the \textit{spectral distribution} of $k$.\\
The kernel function can hence be written as:
\[
k(x,y) = \mathbb{E}_{\omega\sim\Lambda_k}[e^{i\omega^T(x-y)}] = \mathbb{E}_{\omega\sim\Lambda_k}[\cos\left(\omega^T(x-y)\right)] .
\]
By the Monte-Carlo principle, using a sample $(\omega_i)_{i=1}^{D/2}$ i.i.d. from $\Lambda_k$, the feature map $\Phi \colon \mathcal{X}
\rightarrow \mathbb{R}^D$ defined by
\begin{equation}
\label{eq:rff}
\Phi(x) = \sqrt{\frac{2}{D}} 
        \left[\cos(\omega_i^Tx), \ \sin(\omega_i^Tx)\right]_{i=1}^{D/2}
\end{equation}
is such that
$
k(x,y) = \mathbb{E}[\tilde{\Phi}(x)^T\tilde{\Phi}(y)] ,
$
where the expectation is taken with respect to $(\omega_i)_{i=1}^{D/2}$.

Another vectorisation used in practice is 
\begin{equation}
\label{eq:rff2}
\Phi(x) = \sqrt{\frac{2}{D}} 
        \left[\cos(\omega_i^Tx + b_i)\right]_{i=1}^{D},
\end{equation}
where $b_i$ are i.i.d samples from the uniform distribution on $[0, 2\pi]$. See \cite{gundersenRandomFourierFeatures2019} for the detailed computation. Even, if this vectorisation is popular in practice, we would like to point out that the second version yields worst results both in term of variance and upper-bound for the Gaussian kernel \cite{sutherland2015error}.\\

\newpage

\section{Additional figures}

\begin{figure}[ht]
\vskip 0.2in
\begin{center}
\centerline{\includegraphics[width=\columnwidth]{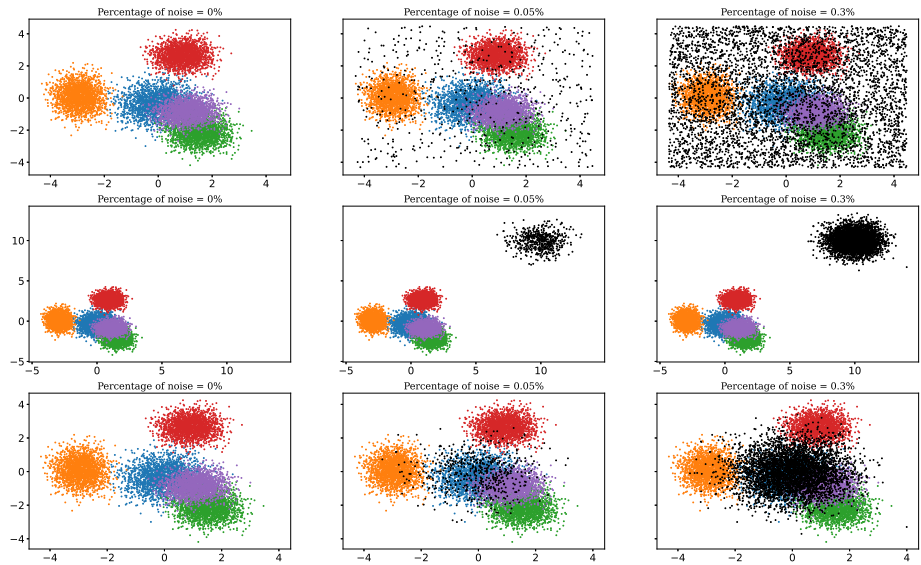}}
\caption{The first row represents the background uniform noise for different values of $\epsilon$. The second row represents the case where a new class appears far from the other distributions. Finally the last row is the scenario where the new class appears close to the other distributions.} 
\label{fig:all noises}
\end{center}
\vskip -0.2in
\end{figure}

Figure \ref{fig:all noises} shows the three types of contamination tested in Section 4.2. 

\end{document}